\documentclass[letterpaper, 10 pt, journal, twoside]{IEEEtran}

\usepackage[numbers,sort&compress]{natbib}
\usepackage{algorithm}
\usepackage{algpseudocode}
\makeatletter
\let\OldStatex\Statex
\renewcommand{\Statex}[1][0]{%
  \setlength\@tempdima{\algorithmicindent}%
  \OldStatex\hskip\dimexpr#1\@tempdima\relax}
\makeatother

\algnewcommand\algorithmicinput{\textbf{Input:}}
\algnewcommand\Input{\item[\algorithmicinput]}
\algnewcommand\algorithmicoutput{\textbf{Output:}}
\algnewcommand\Output{\item[\algorithmicoutput]}

\usepackage{amsmath, amssymb, amsthm}

\usepackage{bbm}  %The \mathbb AMS package only provides blackboard bold capital letters, so we additionally load the \mathbbm{} fonts supplied by the bbm package for e.g. blackboard bolded 1 (vector of ones)
\usepackage{accents}  %To get underbar symbol

\newcommand{\R}{\mathbb{R}}
\newcommand{\bigO}{\mathcal{O}}

\def \Sym{\mathbb{S}}
\def \PSD{\Sym_{+}}
\DeclareMathOperator{\tr}{tr}
\DeclareMathOperator{\Span}{span}
\DeclareMathOperator{\rank}{rank}
\DeclareMathOperator{\Diag}{Diag}
\def \transpose{^\mathsf{T}}
\def \inv{^{-1}}
\def \tinv{^{-\mathsf{T}}}
\DeclareMathOperator{\GL}{GL}
\DeclareMathOperator*{\argmin}{argmin}

\newcommand{\st}{\textnormal{s.t.\; }}
\DeclareMathOperator{\Uniform}{\mathcal{U}}
\def \Nodes {\mathcal{V}}  % Set of nodes
\def \Edges {\mathcal{E}}  % Set of undirected edges
\def \Weights{\mathcal{W}} % Set of edge weights
\newtheorem{thm}{Theorem}

\theoremstyle{definition}

\newtheorem{problem}{Problem}
\newtheorem{remark}{Remark}

\usepackage{xcolor}

\usepackage{epstopdf}
\usepackage{epsfig}
\usepackage{subfigure}

\newcommand{\titlestring}{ {Accelerating Certifiable Estimation with Preconditioned Eigensolvers}}
\newcommand{\authorstring}{David M.\ Rosen}

\usepackage[bookmarks=true, bookmarksnumbered=true, pdfpagemode=UseOutlines, pdfstartview=FitH, 
colorlinks, linkcolor=black, citecolor=black, urlcolor=black, pdftitle={\titlestring}, pdfauthor={\authorstring}]{hyperref}

\newcommand{\algsmall}[1]{{#1}_{\textnormal{min}}}  % Decorator for an estimate of a random value

\newcommand{\edit}[1]{{#1}}

\def \cert{S}

\def \A{\mathcal{A}}

\begin{document}

%Title page
\title{Accelerating Certifiable Estimation with Preconditioned Eigensolvers
\thanks{Manuscript received 18 August 2022; accepted 18 October 2022.} %Use only for final RAL version
\thanks{This letter was recommended for publication by Associate Editor A.~Quattrini Li and Editor L.~Pallottino upon evaluation of the reviewers’ comments.}

\thanks{The author is with the Departments of Electrical \& Computer Engineering and Mathematics, Northeastern University, Boston, MA 02115 USA (email: d.rosen@northeastern.edu).}

\thanks{Digital Object Identifier 10.1109/LRA.2022.3220154}
}%

\author{David M.\ Rosen}

\markboth{IEEE Robotics and Automation Letters. Preprint Version. Accepted Oct 2022}
{Rosen: Accelerating Certifiable Estimation with Preconditioned Eigensolvers}

\maketitle

\begin{abstract}

Convex (specifically semidefinite) relaxation provides a powerful approach to constructing robust machine perception systems, enabling the recovery of \emph{certifiably globally optimal} solutions of challenging estimation problems in many practical settings.  However, solving the large-scale semidefinite relaxations underpinning this approach remains a formidable computational challenge. A dominant cost in many state-of-the-art (Burer-Monteiro factorization-based) certifiable estimation methods is \emph{solution verification} (testing the global optimality of a given candidate solution), which entails computing a minimum eigenpair of a certain symmetric \emph{certificate matrix}.  In this letter, we show how to significantly accelerate this verification step, and thereby the overall speed of certifiable estimation methods.  First, we show that the certificate matrices arising in the Burer-Monteiro approach generically possess spectra that make the verification problem expensive to solve using standard iterative eigenvalue methods.  We then show how to address this challenge using \emph{preconditioned} eigensolvers; specifically, we design a specialized solution verification algorithm based upon the \emph{locally optimal block preconditioned conjugate gradient} (LOBPCG) method together with a simple yet highly effective algebraic preconditioner.  Experimental evaluation on a variety of simulated and real-world examples shows that our proposed verification scheme is very effective in practice, accelerating solution verification by up to \edit{280x}, and the overall Burer-Monteiro method by up to \edit{16x}, versus the standard Lanczos method when applied to relaxations derived from large-scale SLAM benchmarks.
\end{abstract}


\section{Introduction}

\IEEEPARstart{M}{any} fundamental machine perception tasks require the solution of a high-dimensional nonconvex estimation problem; this class includes (for example) the fundamental problems of simultaneous localization and mapping (SLAM) in robotics, 3D reconstruction (in computer vision), and sensor network localization (in distributed sensing), among others.  Such problems are known to be computationally hard to solve in general \cite{Rosen2021Advances}, with many local minima that can entrap the smooth local optimization methods commonly applied to solve them \cite{Dellaert2017Factor}.  The result is that traditional machine perception algorithms (based upon \emph{local} optimization) can be surprisingly brittle, often returning egregiously wrong answers even when the problem to which they are applied is well-posed.

Recent work has shown that \emph{convex} (specifically \emph{semidefinite}) \emph{relaxation} provides a powerful alternative to (heuristic) local search for building highly-reliable machine perception systems, enabling the recovery of \emph{provably globally optimal} solutions of generally-intractable estimation problems in many practical settings (cf.\  \cite{Rosen2021Advances} generally).  However, despite the great promise of this approach, solving the resulting semidefinite relaxations remains a significant computational challenge \cite{Majumdar2019Scalability}. 

One of the most successful strategies to date for solving the large-scale semidefinite relaxations arising in machine perception applications is \emph{Burer-Monteiro factorization} \cite{Burer2003Nonlinear}.  In brief, this method searches for a solution to a (high-dimensional) semidefinite program by alternately applying fast \emph{local} optimization to a surrogate {nonconvex} (but low-dimensional) problem, and then testing whether the recovered critical point is a \emph{global} optimum for the original SDP (which entails calculating a minimum eigenpair of a certain symmetric \emph{certificate matrix} [cf.\ Sec.\ \ref{Burer_Monteiro_section}]).  Surprisingly, \emph{verifying} a candidate solution is frequently the dominant cost in the Burer-Monteiro approach, often requiring an order of magnitude more computational effort (using standard iterative eigenvalue methods) than the entire nonlinear optimization required to \emph{produce} it.

In this letter we show how to significantly accelerate this rate-limiting verification step, and thereby the overall speed of current state-of-the-art certifiable estimation methods.  First, we show that the certificate matrices arising in the Burer-Monteiro approach generically possess spectra that make it expensive to compute a minimum eigenpair using standard iterative eigenvalue methods.   Next, we show how to address this challenge using \emph{preconditioned} eigensolvers; specifically, we design a specialized verification algorithm based upon the \emph{locally optimal block preconditioned conjugate gradient} (LOBPCG) \cite{Knyazev2001LOBPCG} method together with a simple yet highly effective algebraic preconditioner.  Experimental evaluation on a variety of synthetic and real-world examples demonstrates that our proposed verification approach is very effective in practice, accelerating solution verification by up to \edit{280x}, and the overall Burer-Monteiro method by up to \edit{16x}, versus a standard Lanczos method when applied to relaxations derived from large-scale SLAM benchmarks.

\section{Semidefinite optimization using the Burer-Monteiro method}
\label{Burer_Monteiro_section}

In this section we briefly review the Burer-Monteiro method for semidefinite optimization \cite{Burer2003Nonlinear}, which forms the algorithmic foundation of many state-of-the-art certifiable estimation methods (cf.\ \cite{Rosen2021Advances} and the references therein).  We begin with a brief review of semidefinite programs and a discussion of their computational cost.  We then describe the Burer-Monteiro method, which provides a scalable approach to solving SDPs with \emph{low-rank} solutions.  Finally, we discuss the \emph{verification problem} (Problem \ref{verification_problem}) arising in the Burer-Monteiro approach, and identify a key feature that makes this problem expensive to solve using standard iterative eigenvalue methods.

\subsection{Semidefinite programs}
\label{Semidefinite_programs__review_subsection}

Recall that a \emph{semidefinite program} (SDP) is a convex optimization problem of the form \cite{Boyd2004Convex}:
 \begin{equation}
\label{semidefinite_program}
\begin{split}
\min_{X \in \PSD^n} \tr(CX) \quad  \st \A(X) = b,
\end{split}
\end{equation}
where $C \in \Sym^n$, $b \in \R^m$, and $\A \colon \Sym^n \to \R^{m}$ is a linear map:
\begin{equation}
\label{linear_operator_definition}
\A(X)_i \triangleq \tr(A_iX) \quad \forall i \in [m]
\end{equation}
parameterized by coefficient matrices $\lbrace A_i \rbrace_{i = 1}^{m} \subset \Sym^n$.  Problem \eqref{semidefinite_program} can in principle be solved efficiently (i.e.\ in polynomial time) using general-purpose (e.g.\ interior point) optimization methods \cite{Boyd2004Convex,Nocedal2006Numerical}.  In practice, however, the high computational cost of storing and manipulating the $O(n^2)$ elements of the decision variable $X$ prevents such straightforward approaches from scaling effectively to problems in which $n$ is larger than a few thousand \cite{Majumdar2019Scalability}.  Unfortunately, typical instances of many important machine perception problems (e.g.\ SLAM) are often several orders of magnitude larger, placing them well beyond the reach of such general-purpose techniques.

\subsection{Burer-Monteiro factorization}

While problem \eqref{semidefinite_program} is expensive to solve \emph{in general}, the \emph{specific instances} of \eqref{semidefinite_program} arising as relaxations in certifiable estimation often admit \emph{low-rank} solutions.  Any such solution $X^* \in \PSD^n$ can be represented concisely as $X^* = {Y^*}{Y^*}\transpose$, where $Y^* \in \R^{n \times r}$ and $r = \rank X^* \ll n$.

In their seminal work, \citet{Burer2003Nonlinear} proposed to exploit the existence of such low-rank solutions by replacing the high-dimensional decision variable $X$ in \eqref{semidefinite_program} with its rank-$r$ symmetric factorization $YY\transpose$, producing the following (nonconvex) \emph{rank-restricted} version of the problem:
\begin{equation}
\label{low_rank_program}
\begin{split}
\min_{Y \in \R^{n \times r}} \tr(CYY\transpose) \quad \st \A(YY\transpose) = b.
\end{split}
\end{equation}
This substitution has the effect of dramatically reducing the dimensionality of the search space of \eqref{low_rank_program} versus \eqref{semidefinite_program}, as well as rendering the positive-semidefiniteness constraint in \eqref{semidefinite_program} \emph{redundant} (since $YY\transpose \succeq 0$ by construction).  \citet{Burer2003Nonlinear} thus proposed to search for minimizers $X^*$ of \eqref{semidefinite_program} by attempting to recover their low-rank factors $Y^*$ from \eqref{low_rank_program} using fast \emph{local} nonlinear programming methods  \cite{Nocedal2006Numerical}.

\subsection{Ensuring global optimality}
\label{Ensuring_optimality_subsection}
Since the Burer-Monteiro approach entails searching for a \emph{global} minimizer $Y^*$ of the nonconvex problem \eqref{low_rank_program}, one might naturally wonder whether it is possible to \emph{find} such global optima in practice.  Remarkably, recent work has shown that this can in fact be done under quite general conditions, despite problem \eqref{low_rank_program}'s nonconvexity \cite{Boumal2106Nonconvex,Cifuentes2019Polynomial}.  

Suppose that we have applied a numerical optimization method to an instance of problem \eqref{low_rank_program}, and recovered a first-order stationary point $Y \in \R^{n \times r}$. Then either $X = YY\transpose$ is a global minimizer of \eqref{semidefinite_program} [which we can determine by checking the necessary and sufficient KKT conditions for the convex program \eqref{semidefinite_program}], or $X$ is suboptimal, in which case we would like to identify a direction of improvement for the low-rank factor $Y$.  The following result establishes that checking the optimality of $X = YY\transpose$, and constructing a direction of improvement (if necessary), can \emph{both} be accomplished with a \emph{single} minimum-eigenpair calculation:

\begin{thm}[Theorem 4 of \cite{Rosen2020Scalable}]
\label{theorem_of_alternative_for_KKT_points}
Assume that the semidefinite program \eqref{semidefinite_program} satisfies Slater's condition.  Let $Y \in \R^{n \times r}$ be a KKT point of \eqref{low_rank_program} that satisfies the linear independence constraint qualification, $\lambda \in \R^m$ the corresponding Lagrange multiplier, and define:
\begin{equation}
\label{KKT_certificate_matrix_definition}
\cert \triangleq C + \sum_{i = 1}^m \lambda_i A_i.
\end{equation}
Then exactly one of the following two cases holds:
\begin{itemize}
 \item [$(i)$] $\cert \succeq 0$, and $X = YY\transpose$ is a global minimizer of \eqref{semidefinite_program}. 
 \item [$(ii)$]  There exists $v \in \R^n$ such that $v\transpose \cert v < 0$, and in that case, $Y_{+} = \begin{pmatrix} Y & 0 \end{pmatrix} \in \R^{n \times (r+1)}$ is a KKT point of \eqref{low_rank_program} attaining the same objective value as $Y$, and $\dot{Y}_{+} = \begin{pmatrix} 0 & v \end{pmatrix} \in \R^{n \times (r+1)}$ is a feasible second-order direction of descent from $Y_{+}$.
\end{itemize}
\end{thm}

\noindent In brief, Theorem \ref{theorem_of_alternative_for_KKT_points} establishes that the global optimality of $X = YY\transpose$ as a solution of \eqref{semidefinite_program} can be verified by checking whether the \emph{certificate matrix $\cert$} defined in \eqref{KKT_certificate_matrix_definition} is positive semidefinite, and in the event that it is \emph{not}, that any direction of negative curvature $v \in \R^n$ of $S$ provides a direction of improvement $\dot{Y}_{+}$ from the embedding $Y_{+}$ of the current critical point $Y$ into a higher-dimensional instance of \eqref{low_rank_program}.   

This theorem of the alternative suggests a natural algorithm for recovering minimizers $X^*$ of \eqref{semidefinite_program} via a \emph{sequence} of local optimizations of \eqref{low_rank_program} \cite{Rosen2020Scalable,Boumal2015Riemannian,Journee2010LowRank}.  Starting at some (small) initial rank $r$, apply local optimization to \eqref{low_rank_program} to obtain a stationary point $Y$.  If $X = YY\transpose$ is optimal for \eqref{semidefinite_program}, then return $Y$; otherwise, increment the rank parameter $r$, and restart the local optimization for the next instance of \eqref{low_rank_program} using the direction of improvement $\dot{Y}_{+}$ provided by Theorem \ref{theorem_of_alternative_for_KKT_points}.

\subsection{The solution verification problem}
\label{verification_problem_subsection}

Surprisingly, simply \emph{verifying} the optimality of a candidate solution $X = YY\transpose$ (by calculating a minimum eigenpair of the certificate matrix $\cert$ in \eqref{KKT_certificate_matrix_definition} and applying Theorem \ref{theorem_of_alternative_for_KKT_points}) is frequently the dominant cost in the Burer-Monteiro approach, often requiring an order of magnitude more computational effort than the {entire nonlinear optimization} \eqref{low_rank_program} required to \emph{produce} $Y$ itself (cf.~Sec.~\ref{Experimental_results_section}).  This turns out to be a consequence of a particular feature of the spectra of the certificate matrices $\cert$ arising in the Burer-Monteiro method.

\begin{remark}[The spectrum of the certificate matrix]
\label{computational_challenge_in_solution_verification_remark}
A straightforward calculation shows that the first-order stationarity conditions for \eqref{low_rank_program} can be expressed as (cf.~\cite[Thm.~2]{Rosen2020Scalable}):
\begin{equation}
\label{first_order_stationarity_conditions_for_Burer_Monteiro_problems}
\cert Y = 0.
\end{equation}
This implies that $\cert$ will have $0$ as an eigenvalue of multiplicity \emph{at least} $\rank Y$; that is, $\cert $ always has a \emph{cluster} of eigenvalues at $0$.  It is this eigenvalue clustering that makes calculating a minimum eigenpair of $\cert $ expensive.  In detail, if $\cert \succeq 0$, then $\algsmall{\lambda}(\cert) = 0$, so the target eigenvalue is embedded in a tight cluster, while if $\algsmall{\lambda}(\cert) < 0$ but has small magnitude, it is difficult to estimate accurately because the cluster of $0$ eigenvalues nearby ensures that the relevant eigenvalue gap $\gamma \le \lvert \algsmall{\lambda}(\cert) - 0 \rvert = \lvert \algsmall{\lambda}(\cert) \rvert$ is small.  In effect, equation \eqref{first_order_stationarity_conditions_for_Burer_Monteiro_problems} guarantees that the minimum-eigenpair calculation is almost always poorly-conditioned, and therefore expensive to solve using standard iterative eigenvalue methods (cf.\ Sec.\ \ref{related_work_section}). 
\end{remark}

In light of Theorem \ref{theorem_of_alternative_for_KKT_points} and Remark \ref{computational_challenge_in_solution_verification_remark}, we are thus interested in developing more computationally efficient methods of solving the following (\emph{solution}) \emph{verification problem}:

\begin{problem}[Verification problem]
\label{verification_problem}
Given a matrix $\cert \in \Sym^n$, determine whether $\cert \succeq 0$, and if it is not, calculate a vector $x \in \R^n$ such that $x\transpose \cert x < 0$.
\end{problem}

\section{Related work}
\label{related_work_section}

In this section we review prior work pertaining to the solution of the verification problem (Problem \ref{verification_problem}) from two areas: algorithms for the symmetric eigenvalue problem (Section \ref{Related_work_subsection_on_symmetric_eigenvalue_methods}), and certifiable estimation (Section \ref{Related_work_subsection_on_certifiable_estimation}).

\subsection{Algorithms for the symmetric eigenvalue problem}
\label{Related_work_subsection_on_symmetric_eigenvalue_methods}

Standard large-scale eigenvalue algorithms are \emph{iterative projection methods}, meaning that they recover approximate eigenpairs of a matrix $A \in \Sym^n$ by searching for approximate eigenvectors within a (computationally tractable) low-dimensional {subspace} of $\R^n$ \cite[Chp.~3]{Bai2000Templates}.  This  includes the \emph{power} and \emph{subspace iteration} methods (for computing \emph{dominant} eigenpairs), and the \emph{Lanczos method}, which is the preferred technique for computing a small number of \emph{extremal} eigenpairs of $A$ \cite[Chp.~4]{Bai2000Templates}.    An important property of these methods is that their convergence rates depend crucially upon the \emph{distribution} of $A$'s eigenvalues, and in particular how \emph{well-separated} a target eigenvalue is from the remainder of $A$'s spectrum (relative to $A$'s spectral norm).  For example, letting $\lambda_1 \le \lambda_2 \le \dotsb \le \lambda_K$ denote $A$'s {distinct} eigenvalues, the number of iterations required to estimate a minimum eigenvector $x_1$ of $A$ using the Lanczos method is approximately $\bigO(\sqrt{\lVert A \rVert / \gamma })$, where $\gamma \triangleq \lambda_2 - \lambda_1$ is the \emph{spectral gap} between $A$'s minimum eigenvalue $\lambda_1$ and the next-largest eigenvalue $\lambda_2$ (cf.\ \cite[Thm.~6.3]{Saad2011Eigenvalue} and the discussion in \cite[Sec.\ 6.1]{Tian2021Distributed}).  Consequently, computing a minimum eigenpair of $A$ using the Lanczos method is expensive when $\gamma \ll \rVert A \rVert$.  Unfortunately, as shown in Section \ref{verification_problem_subsection}, this case generically holds for the verification problems arising in the Burer-Monteiro approach.

Given the sensitivity of iterative eigenvalue methods to the distribution of $A$'s spectrum, a natural idea for improving computational performance is to precondition the eigenvalue problem by applying a \emph{spectral transformation} that sends the desired eigenvalue $\lambda$ of $A$ to a well-separated \emph{extremal} eigenvalue \cite[Sec.~3.3]{Bai2000Templates}.  The most common such method is the \emph{shift-and-invert} (SI) spectral transformation, in which the matrix $A$ is replaced by the resolvent $R_{\sigma} \triangleq (A - \sigma I)\inv$ for $\sigma \notin \Lambda(A)$.  This transformation maps each eigenpair $(\lambda, x)$ of $A$ to the eigenpair $(\nu, x)$ of $R_{\sigma}$, where $\nu = (\lambda - \sigma)\inv$; in particular, it sends the eigenvalues of $A$ nearest to the spectral shift $\sigma$ to well-separated extremal eigenvalues of $R_{\sigma}$, thus enabling their efficient recovery.  Unfortunately, basic SI is insufficient to precondition the verification problem, since we do not know \emph{a priori} where the desired minimum eigenvalue $\algsmall{\lambda}(\cert)$ is (indeed, determining this is part of the problem!)

Some iterative eigenvalue methods employ what can be interpreted as \emph{adaptive} SI preconditioning schemes.  In brief, these methods leverage the updated eigenvalue estimates $\theta_k$ calculated in each iteration to construct an improved SI preconditioner $R_{\theta_k}$ for the \emph{next} iteration; well-known examples of this approach include \emph{Rayleigh quotient iteration} (RQI) \cite[Sec.\ 4.3.3]{Bai2000Templates} and the \emph{Jacobi-Davidson} (JD) method \cite[Sec.\ 4.7]{Bai2000Templates}, which can be interpreted as a kind of inexact subspace-accelerated RQI method. While these techniques promote rapid convergence, they require solving sequences of linear systems with \emph{varying} coefficient matrices $A - \theta_k I$, which in practice entails either repeated matrix factorizations or the repeated construction of preconditioners to use with iterative linear system solvers (e.g.\ MINRES), both of which can be dominant costs in large-scale calculations.

In light of these considerations, we propose to adopt the \emph{locally optimal block preconditioned conjugate gradient} (LOBPCG) method \cite{Knyazev2001LOBPCG} as the basis of our fast verification approach.  In brief, LOBPCG is a preconditioned subspace-accelerated simultaneous iteration scheme for recovering a few \emph{algebraically-smallest} eigenpairs of $A$.  An important distinguishing feature of this method is that it employs a single (\emph{constant}) preconditioner, which enables preconditioning the verification problem with modest computational overhead.

\subsection{Certifiable estimation}
\label{Related_work_subsection_on_certifiable_estimation}
In consequence of the increasing interest in certifiable estimation methods (cf.\ \cite{Rosen2021Advances} and the references therein), several recent works have made use of (specific instances of) Theorem \ref{theorem_of_alternative_for_KKT_points} to certify the global optimality of candidate solutions to machine perception problems, whether as part of a Burer-Monteiro method \cite{Rosen2019SESync,Rosen2020Scalable,Briales2017Cartan,Fan2019Efficient,Tian2021Distributed,Dellaert2020Shonan} or as a standalone procedure applied to estimates recovered from more traditional heuristic local search \cite{Carlone2015Lagrangian,Carlone2015Duality,Briales2016Fast}.  These prior works solve the verification problem by applying Krylov subspace methods (typically the Lanczos method) directly to the certificate matrix $\cert$ to recover a minimum eigenpair.  Reference \cite{Rosen2017Computational} proposed a simple spectral-shifting strategy to convert the verification problem to the calculation of a \emph{dominant} eigenpair (for which the Lanczos method exhibited better empirical performance), while \cite{Tian2021Distributed} proposed to employ \emph{accelerated} power iterations, which are more amenable to a distributed implementation.

To the best of our knowledge, this letter is the first to investigate the design of specialized algorithms for solving Problem \ref{verification_problem}, and in particular the first to propose the use of \emph{preconditioned} eigensolvers to ameliorate the structural ill-conditioning identified in Section \ref{verification_problem_subsection} whenever the calculation of a vector $x \in \R^n$ satisfying $x\transpose \cert x < 0$ is required.

\section{Accelerating solution verification with preconditioned eigensolvers}
\label{Fast_verification_method_section}
In this section we present our proposed fast verification scheme. We begin in Sec.~\ref{LOBPCG_method_section} with a review of the LOBPCG method, which forms the basis of our approach, and highlight some of the features that make it especially well-suited to solving the verification problem.  We then describe our  preconditioning strategy for use with LOBPCG in Sec.~\ref{modified_incomplete_symmetric_LDL_preconditioning}.  Finally, we present our complete algorithm in Sec.~\ref{Fast_verification_algorithm_subsection}.

\subsection{The LOBPCG eigensolver}
\label{LOBPCG_method_section}

The \emph{locally optimal block preconditioned conjugate gradient} (LOBPCG) method \cite{Knyazev2001LOBPCG,Knyazev2007BLOPEX} is an iterative eigenvalue method for calculating the $k$ algebraically-smallest eigenpairs $(\lambda, x)$ of the generalized symmetric eigenvalue problem:
\begin{equation}
 \label{generalized_symmetric_eigenvalue_problem}
 Ax = \lambda Bx
\end{equation}
where $A, B \in \Sym^n$ and $B \succ 0$.  This method exploits the Courant-Fischer variational characterization of a minimum eigenpair $(\lambda, x)$ of \eqref{generalized_symmetric_eigenvalue_problem} as a solution of the following minimization problem \cite[Sec.\ 9.2.6]{Saad2011Eigenvalue}:
\begin{equation}
\label{Rayleigh_quotient_optimization_for_generalized_symmetric_eigenvalue_problem}
\lambda = \min_{x \in \R^n}x\transpose A x \quad \st \quad x\transpose Bx = 1
\end{equation}
to search for $x$ by applying first-order optimization to \eqref{Rayleigh_quotient_optimization_for_generalized_symmetric_eigenvalue_problem}.
 
In detail, the Lagrangian $L \colon \R^{n} \times \R \to \R$ of \eqref{Rayleigh_quotient_optimization_for_generalized_symmetric_eigenvalue_problem} is:
\begin{equation}
\begin{gathered}
L(x, \lambda) \triangleq x\transpose A x - \lambda (x\transpose B x - 1)
\end{gathered}
\end{equation}
and a straightforward calculation shows that the gradient of $L(x, \lambda)$ with respect to $x$ is:
\begin{equation}
\label{gradient_of_Lagrangian}
 \nabla_x L(x, \lambda) = 2Ax - 2\lambda Bx.
\end{equation}
Comparing \eqref{generalized_symmetric_eigenvalue_problem} and \eqref{gradient_of_Lagrangian} reveals that the KKT points $(x, \lambda)$ of \eqref{Rayleigh_quotient_optimization_for_generalized_symmetric_eigenvalue_problem} are precisely the eigenpairs of \eqref{generalized_symmetric_eigenvalue_problem}; furthermore, given any \emph{estimate} $(\theta, x)$ for a minimum eigenpair, we see that the residual $r$ of the generalized eigenvalue equation \eqref{generalized_symmetric_eigenvalue_problem}:
\begin{equation}
\label{residual_of_pair_estimate}
r \triangleq Ax - \theta Bx
\end{equation}
is parallel to the gradient $\nabla_x L(x, \theta)$ in \eqref{gradient_of_Lagrangian}.  A basic first-order optimization method would thus take $-r$ as a search direction to calculate an improved minimum eigenvector estimate \cite{Nocedal2006Numerical}.

LOBPCG improves upon a basic first-order optimization approach in three important ways.  First and foremost, it exploits the fact that the objective in \eqref{Rayleigh_quotient_optimization_for_generalized_symmetric_eigenvalue_problem} is a quadratic parameterized by $A$ by replacing the residual (gradient) search direction $r$ in \eqref{residual_of_pair_estimate} with the \emph{preconditioned} gradient search direction:
\begin{equation}
\label{preconditioned_gradient_search_direction}
w \triangleq Tr,
\end{equation}
where $T \in \Sym^n_{++}$ is a user-defined preconditioning operator.  This should be chosen to approximate $A\inv$ in the sense that $\kappa(TA) \ll \kappa(A)$; this has the effect of ``undistorting'' the contours of the quadratic objective in \eqref{Rayleigh_quotient_optimization_for_generalized_symmetric_eigenvalue_problem}, producing significantly higher-quality search directions.  This strategy has two major advantages versus the alternative SI preconditioning strategy used in e.g.\ Rayleigh quotient iteration and the Jacobi-Davidson method.  First, the figure of merit  $\kappa(TA)$ for the preconditioner $T$ is the same one appearing in the design of preconditioners for iterative linear system solvers (such as MINRES), which enables the direct application of this large body of prior work to the generalized eigenvalue problem \eqref{generalized_symmetric_eigenvalue_problem}.  Second, in contrast to the \emph{varying} resolvents $R_{\theta_k} \triangleq (A - \theta_k I)\inv$ used in adaptive SI preconditioning, the preconditioner $T$ used in LOBPCG is designed for the parameter matrix $A$ appearing in \eqref{Rayleigh_quotient_optimization_for_generalized_symmetric_eigenvalue_problem}, and is therefore \emph{constant}.  Since constructing a preconditioner is often done via incomplete matrix factorization (which is expensive at large scale), enabling the use of a constant preconditioner $T$ can provide substantial computational savings.

Second, LOBPCG employs \emph{subspace acceleration}: instead of searching \emph{solely} over the 1-dimensional subspace spanned by $w$ (as in basic preconditioned gradient descent), at each iteration it calculates the next eigenvector estimate $x_{i+1}$ as a linear combination of the preconditioned gradient direction $w_{i}$, the current iterate $x_{i}$, \emph{and} the previous iterate $x_{i-1}$:
\begin{equation}
\label{LOBPCG_update_equation}
x_{i+1} = \alpha_{i} w_{i} + \beta_{i} x_{i} + \gamma_{i} x_{i-1},
\end{equation}
where $\alpha_{i}, \beta_{i}, \gamma_{i} \in \R$ are coefficients.  The use of a larger (3-dimensional) search space in \eqref{LOBPCG_update_equation} provides more flexibility in the choice of update $x_{i+1}$, and therefore the opportunity for enhanced performance.  In particular, we remark that standard accelerated gradient methods (such as Nestorov, heavy ball, and the accelerated power method) all correspond to special cases of \eqref{LOBPCG_update_equation} for specific choices of  $\alpha_{i}$, $\beta_{i}$, and $\gamma_{i}$.  Moreover, LOBPCG further improves upon standard accelerated gradient schemes by exploiting the specific form of \eqref{Rayleigh_quotient_optimization_for_generalized_symmetric_eigenvalue_problem} to dynamically compute \emph{optimal} choices of $\alpha_{i}$, $\beta_{i}$, and $\gamma_{i}$ in \eqref{LOBPCG_update_equation} in each iteration.\footnote{This is the sense in which LOBPCG is ``locally optimal''.} Specifically, the updated estimate $x_{i+1}$ is chosen to minimize the Rayleigh quotient over the 3-dimensional subspace spanned by $w_{i}$, $x_{i}$, and $x_{i-1}$:
\begin{equation}
\label{LOBPCG_update_step}
x_{i+1} \in \argmin_{x \in \Span \left\lbrace w_{i},\: x_{i}, \: x_{i-1} \right \rbrace} x\transpose A x \quad  \st \quad x\transpose Bx = 1.
\end{equation}
In practice, the minimization \eqref{LOBPCG_update_step} amounts to solving a projected generalized eigenvalue problem on the 3-dimensional subspace $\mathcal{S} = \Span \lbrace w_{i}, \: x_{i}, \: x_{i-1} \rbrace$, which can be done simply and efficiently using the Rayleigh-Ritz procedure (Alg.~\ref{RayleighRitz_algorithm}).

Finally, LOBPCG extends the single-vector iteration described in \eqref{residual_of_pair_estimate}--\eqref{LOBPCG_update_step} to \emph{simultaneous iteration} on a \emph{block} $X \in \R^{n \times m}$ of $m$ eigenvector estimates.  This provides two major advantages.  First, it permits the \emph{simultaneous} recovery of the $k \le m$ algebraically-smallest eigenpairs of \eqref{generalized_symmetric_eigenvalue_problem}.  Second, a rough analysis suggests (and numerical experience confirms) that the convergence rate of the $k$ smallest eigenpairs in a  block   of size $m$ depends upon the eigengap $\gamma_{m,k} \triangleq \lambda_{m+1} - \lambda_k$ between the largest desired eigenvalue $\lambda_k$ and the smallest eigenvalue $\lambda_{m+1}$ \emph{not} included in the block (cf.\ the discussion around eq.~(5.5) in \cite{Knyazev2001LOBPCG}).  It is therefore advantageous to choose the block size $m$ greater than the number of desired eigenpairs $k$ if doing so will significantly enhance the eigengap $\gamma_{m,k}$.  This capability is especially useful for addressing clustered eigenvalues, which we have seen  are the primary challenge in solving the verification problem.

The complete LOBPCG algorithm is shown as Algorithm \ref{LOBPCG_algorithm} (cf.\ Algorithms 1 and 2 of \cite{Duersch2018LOBPCG}).

\begin{algorithm}[t]
\caption{The Rayleigh-Ritz procedure}
\label{RayleighRitz_algorithm}
\begin{algorithmic}[1]
\Input  Symmetric linear operators $A, B \in \Sym^n$ with $B \succ 0$, full-rank basis matrix $S \in \R^{n \times k}$.
\Output Diagonal matrix $\Theta = \Diag(\theta_1, \dotsc, \theta_k) \in \Sym^k$ of Ritz values with $\theta_1 \le \dotsb \le \theta_k$, matrix $C \in \GL(k)$ satisfying $C\transpose (S\transpose B S)C = I_k$ and $C\transpose (S\transpose A S)C = \Theta$.
\Function{RayleighRitz}{$A$, $B$, $S$}
\State $D = \Diag(S\transpose BS)^{-1/2}$.
\State Cholesky factorize  $R\transpose R = DS\transpose B SD$.
\State Calculate symmetric eigendecomposition:
\begin{equation*}
 Q\Theta Q\transpose = R\tinv DS\transpose ASD R\inv
\end{equation*}
\Statex[1] with eigenvalues sorted in ascending order.
\State $C = D R\inv Q$.
\State \Return $(\Theta, C)$
\EndFunction
\end{algorithmic}
\end{algorithm}

\begin{algorithm}[t]
\caption{The locally optimal block preconditioned conjugate gradient (LOBPCG) method}
\label{LOBPCG_algorithm}
\begin{algorithmic}[1]
\Input Symmetric linear operators $A, B, T$ with $B, T \succ 0$, initial eigenvector estimates $X \in \R^{n \times m}$, number of desired eigenpairs $k \le m$.
\Output Estimates of the $k$ algebraically-smallest eigenpairs  of the generalized symmetric eigenvalue problem \eqref{generalized_symmetric_eigenvalue_problem}.
\Function{LOBPCG}{$A$, $B$, $T$, $X$, $k$}
\Statex[1] // Initialization
\State $(\Theta, C)$ = \Call{RayleighRitz}{$A$, $B$, $X$}
\State $X = XC$  \Comment{$B$-orthonormalize initial $X$}
\State $R = AX - BX\Theta$  \Comment{Initial residuals}
\State $P = \emptyset$  \Comment{Initialize $P$}
\Repeat{}
\State $W = TR$  \Comment{Precondition residuals}
\State $S = \begin{pmatrix} X & W & P \end{pmatrix}$  \Comment{Search space basis}
\State $(\Theta, C)$ = \Call{RayleighRitz}{$A$, $B$, $S$}
\State $X = SC_{:, 1:m}$  \Comment{Update eigenvector estimates}
\State $R = AX - BX \Theta_{1:m, 1:m}$ \Comment{Compute residuals}
\State $P = S_{:, m+1:end}\: C_{m+1:end, 1:m}$
\State Determine the number $n_c$ of the $k$ algebraically-
\Statex[2] smallest eigenpairs $(\theta_i, x_i$) that have converged.
\Until{$n_c == k$}
\State \Return $(\Theta_{1:k, 1:k}, X_{:, 1:k})$ 
\EndFunction
 \end{algorithmic}
\end{algorithm}

\subsection{Preconditioner design} 
\label{modified_incomplete_symmetric_LDL_preconditioning}

In this section, we propose a simple yet highly effective algebraic preconditioning strategy for use with LOBPCG, based upon a modified incomplete symmetric indefinite factorization.   

Our approach is based upon the following simple argument.  Recall that every nonsingular matrix $A \in \Sym^n$ admits a \emph{Bunch-Kaufman} factorization of the form \cite[Sec.\ 4.4]{Golub1996Matrix}:
\begin{equation}
\label{Bunch_Kaufman_factorization}
PAP\transpose = L\Diag(D_1, \dotsc, D_K)L\transpose,
\end{equation}
where $P$ is a permutation matrix, $L$ is unit lower-triangular, and $D_k$ is a $1 \times 1$ or $2 \times 2$ symmetric nonsingular matrix for all $k \in [K]$.  Given the factorization \eqref{Bunch_Kaufman_factorization}, consider the matrix:
\begin{equation}
\label{T_matrix_definition}
T \triangleq P\transpose L\tinv \Diag\left(D_1^{+}, \dotsc, D_K^{+}\right)L\inv P,
\end{equation}
where for any nonsingular matrix $M \in \Sym^p$ with symmetric eigendecomposition $M = Q \Diag(\lambda_1, \dotsc, \lambda_p) Q\transpose$, we define:
\begin{equation}
\label{positive_definite_modification_of_symmetric_indefinite_matrix}
M^{+} \triangleq Q\Diag\left(\lvert \lambda_1\rvert\inv, \dotsc, \lvert \lambda_p \rvert\inv\right)Q\transpose \in \Sym_{++}^p
\end{equation}
to be the positive-definite matrix obtained from $M$ by replacing each of $M$'s eigenvalues with the reciprocal of its absolute value.  Then $T \succ 0$, and a straightforward calculation yields:
\begin{equation}
\label{TA_product}
 TA = P\inv L\tinv \Diag\left(D_1^{+} D_1, \dotsc, D_K^{+} D_K \right) L\transpose P.
\end{equation}
Equation \eqref{TA_product} shows that the product $TA$ is similar to a block-diagonal matrix whose blocks $D_k^{+}D_k$ all have eigenvalues contained in $\lbrace \pm 1\rbrace$.  This proves that $\Lambda(TA) \subseteq \lbrace \pm 1 \rbrace$, and therefore $\kappa(TA)= 1$; that is, the matrix $T$ defined in \eqref{T_matrix_definition} attains the \emph{smallest possible} condition number $\kappa(TA)$, and is therefore an ``ideal'' preconditioner for use with LOBPCG.

Since computing a matrix decomposition can be expensive when $A$ is large, in practice we propose to construct our preconditioners by replacing the \emph{exact} factorization \eqref{Bunch_Kaufman_factorization} with a tractable \emph{approximation} $PAP\transpose \approx \tilde{L} \Diag(\tilde{D}_1, \dotsc, \tilde{D}_K)\tilde{L}\transpose$.  Specifically, we propose to use a \emph{limited-memory incomplete} symmetric indefinite $LDL\transpose$ factorization \cite{Greif2017SymILDL}; this provides direct control over the number of nonzero elements appearing in the unit lower-triangular factors $\tilde{L}$, and therefore scales gracefully (in both memory and computation) to high-dimensional problems.  We construct each diagonal block $\tilde{D}_k^{+}$  by directly computing and modifying the symmetric eigendecomposition of $\tilde{D}_k$ as in \eqref{positive_definite_modification_of_symmetric_indefinite_matrix}; this is again efficient, as each block is at most 2-dimensional.  Finally, we avoid \emph{explicitly instantiating} $T$ as a (generically-dense) matrix.  Instead, whenever we must compute a product with $T$, we sequentially apply each of the linear operators on the right-hand side of \eqref{T_matrix_definition}; this requires only 2 permutations, 2 sparse triangular solves, and a set of parallel multiplications with the (very small) diagonal blocks $\tilde{D}_k^{+}$ per application of $T$, and is thus again efficient.

\subsection{An accelerated solution verification method}
\label{Fast_verification_algorithm_subsection}

Our complete proposed method for solving the verification problem (Problem \ref{verification_problem}) is shown as Algorithm \ref{FastVerification_algorithm}. Our method first tests the positive-semidefiniteness of the regularized certificate matrix $M \triangleq \cert +  \eta I$ (where $\eta > 0$ is a user-selectable regularization parameter) by attempting to compute its Cholesky decomposition.  Note that the use of regularization here is in fact \emph{necessary}: since $0$ is always an eigenvalue of $\cert$ [cf.\ eq.\ \eqref{first_order_stationarity_conditions_for_Burer_Monteiro_problems}], then in the event $\cert \succeq 0$ (corresponding to a \emph{global} minimizer in the Burer-Monteiro method), $\cert$'s algebraically-smallest eigenvalue will be $\algsmall{\lambda}(\cert) = 0$, and therefore the result of \emph{any} numerical procedure for testing $\cert$'s positive-semidefiniteness [i.e.\ the sign of $\algsmall{\lambda}(\cert)$] will be dominated by finite-precision errors.  On the other hand,  if $\cert \succeq 0$ then $M \succ 0$, and therefore the Cholesky factorization attempted in line \ref{M_Cholesky_factorization} exists and can be computed numerically stably \cite[Sec.~4.2.7]{Golub1996Matrix}; moreover, the factorization $M = LL\transpose$ itself then serves as a certificate that $M = \cert + \eta I \succeq 0$, or equivalently that $\cert \succeq - \eta I$.  The parameter $\eta$ thus plays the role of a numerical tolerance for testing $\cert$'s positive-semidefiniteness.

If the Cholesky factorization attempted in line \ref{M_Cholesky_factorization} does \emph{not} succeed, then $\cert \not \succeq 0$, and we must therefore recover a vector $x \in \R^n$ satisfying $x\transpose \cert x < 0$.  Our algorithm achieves this by applying LOBPCG to compute a minimum eigenvector of the regularized certificate matrix $M$, using the modified incomplete factorization-based preconditioning strategy of Sec.~\ref{modified_incomplete_symmetric_LDL_preconditioning}.  Once again we employ $M$  instead of $\cert$ because the former is nonsingular (except in the unlikely case that $-\eta \in \Lambda(\cert)$), so that the factorization and corresponding preconditioner $T$ computed in lines \ref{incomplete_factorization}--\ref{construct_precondtioner} are well-defined.

\begin{algorithm}[t]
\caption{Fast solution verification}
\label{FastVerification_algorithm}
\begin{algorithmic}[1]
\Input Certificate matrix $\cert \in \Sym^n$, minimum eigenvalue numerical nonnegativity tolerance $\eta > 0$, blocksize $m \ge 1$.
\Output Factorization $\cert + \eta I = LL\transpose$ certifying $\cert \succeq - \eta I$, or a Ritz pair $(\lambda, x)$ satisfying $\lambda = x\transpose \cert x$ and $\lambda < 0$.
\Function{FastVerification}{$\cert$, $\eta$, $m$}
\State Construct regularized certificate matrix $M = \cert + \eta I$.
\State Attempt Cholesky factorization of $M$. \label{M_Cholesky_factorization}
\If{$M = LL\transpose$} 
\State \Return Cholesky factor $L$. \Comment{$\cert \succeq -\eta I$}
\EndIf \label{end_of_direct_test_for_positive_semidefiniteness}
\State Incomplete symmetric indefinite factorization:  \label{incomplete_factorization}
\begin{equation*}
PMP\transpose \approx L\Diag(D_1, \dotsc, D_K)L\transpose.
\end{equation*}

\State Compute $D_k^{+}$ via \eqref{positive_definite_modification_of_symmetric_indefinite_matrix}  $\forall k \in [K]$.  \Comment{Correct inertia}
\State Construct preconditioner $T$ as in \eqref{T_matrix_definition}. \label{preconditioner_construction_end} \label{construct_precondtioner}
\State Randomly sample initial block $X \in \R^{n \times m}$.
\State $(\theta, x) = \Call{LOBPCG}{M, I, T, X, 1}$.
\State \Return $(\theta - \eta, x)$
\EndFunction
 \end{algorithmic}
\end{algorithm}

\section{Experimental results}
\label{Experimental_results_section}

In this section we evaluate the performance of our fast verification method (Algorithm \ref{FastVerification_algorithm}) on a variety of synthetic and real-world examples. All experiments are performed on a Lenovo T480 laptop with an Intel Core i7-8650U 1.90 GHz processor and 16 Gb of RAM running Ubuntu 22.04. Our experimental implementation of Algorithm \ref{FastVerification_algorithm} is written in C++, using  CHOLMOD \cite{Chen2009Algorithm} to perform the Cholesky factorization in line \ref{M_Cholesky_factorization}, and SYM-ILDL \cite{Greif2017SymILDL} to compute the incomplete symmetric indefinite factorization in line \ref{incomplete_factorization}.   

\subsection{Simulation studies}
Our first set of experiments is designed to study how the performance of Algorithm \ref{FastVerification_algorithm} depends upon (i) the eigenvalue gap $\gamma$ associated with the minimum eigenvalue $\algsmall{\lambda}(\cert)$  (cf.\ Sec.\ \ref{Related_work_subsection_on_symmetric_eigenvalue_methods}) and (ii) the dimension of $\cert$.  

To that end, we generate a set of suitable test instances using Algorithm \ref{test_matrix_generation_algorithm}.  In brief, this method constructs a matrix $\cert$ by randomly sampling a weighted \emph{geometric random graph} $G$ on the unit square $[0, 1]^2 \subset \R^2$ \cite{Penrose2003Random}, forming its graph Laplacian $L(G)$ \cite{Chung1997Spectral}, and then extending this matrix by a single row and column whose only nonzero value is the final element $-\gamma$ on the diagonal.  This generative procedure is designed to capture several important properties of the certificate matrices arising in real-world instances of Problem \ref{verification_problem}, including:
\begin{itemize}
 \item The sampled matrices are of the form $\cert = L(G) + E$, where $L(G)$ is the Laplacian of a weighted graph $G$ and $E$ is a sparse matrix;
  \item The topology of the graph $G$ captures the (spatial) \emph{locality}-based connectivity typical of real-world measurement networks in robotics and computer vision \cite{Rosen2021Advances};
 \item There is variation in the weights of $G$'s edges (corresponding to the variation of measurement precisions in real-world measurement networks).
\end{itemize}
Finally, since $L(G) \succeq 0$ and $0 \in \Lambda(L(G))$ (as $L(G)$ is a graph Laplacian), it follows from \eqref{construction_of_test_matrix} that $\algsmall{\lambda}(\cert) = -\gamma$ and the eigenvalue gap is $\lvert -\gamma - 0 \rvert = \gamma$; that is, we can directly and \emph{deterministically} control the relevant eigenvalue gap $\gamma$ of the random test matrices $\cert$ sampled from Algorithm \ref{test_matrix_generation_algorithm}.

\begin{algorithm}[t]
\caption{Test matrix sampler}
\label{test_matrix_generation_algorithm}
\begin{algorithmic}[1]
\Function{SampleTestMatrix}{$N$, $r$, $W_{\textnormal{max}}$, $\gamma$}.
\Statex[1] // Sample geometric random graph $G$
\State Sample $N$ vertices distributed uniformly randomly
\Statex[1]  over the unit square: $\Nodes \triangleq \lbrace x_i \sim \Uniform([0,1]^2) \mid i \in [N] \rbrace$.
\State Construct edge set by joining vertices closer than a
\Statex[1] distance $r$: $\Edges \triangleq \lbrace \lbrace i,j \rbrace \mid i \ne j \textnormal{ and } d(x_i, x_j) < r \rbrace$.
\State Sample edge weights uniformly randomly in $[0, W_{\textnormal{max}}]$:
\Statex[1] $\Weights \triangleq \lbrace w_{ij} \sim \Uniform([0, W_{\textnormal{max}}]) \mid \lbrace i,j \rbrace \in \Edges \rbrace$.
\State Construct weighted graph $G \triangleq (\Nodes, \Edges, \Weights)$.
\Statex[1] // Construct test matrix
\State Construct Laplacian matrix $L(G)$ of $G$, and set:
\begin{equation}
\label{construction_of_test_matrix}
\cert \triangleq 
\begin{pmatrix}
L(G) & 0 \\
0 & -\gamma
\end{pmatrix}.
 \end{equation}
\State \Return $\cert$
\EndFunction
 \end{algorithmic}
\end{algorithm}

For the following experiments we implemented Algorithm \ref{test_matrix_generation_algorithm} in Python, using the \texttt{random\_geometric\_graph} method in the NetworkX library to sample the graphs $G$, with default parameters\footnote{This choice of $r$ ensures that the sampled graphs $G$ are asymptotically almost surely connected as $N \to \infty$ \cite[Thm.\ 13.2]{Penrose2003Random}.} $r = 1.25 \sqrt{\log(N) / (\pi N)}$ and $W_{\textnormal{max}} = 10^3$, while varying $N$ and $\gamma$.  As our baseline for comparison, we apply the Lanczos method to compute a minimum eigenpair of $\cert$.  Both methods employ the relative-error convergence criterion:
\begin{equation}
\lVert \cert x - \theta x\rVert \le \tau \lvert \theta \rvert
\end{equation}
for eigenpair estimates $(\theta, x)$, with $\tau = 10^{-2}$.

\begin{figure}
 \subfigure[Computation time vs.\ eigengap $\gamma$ ]{\includegraphics[width=.495\columnwidth]{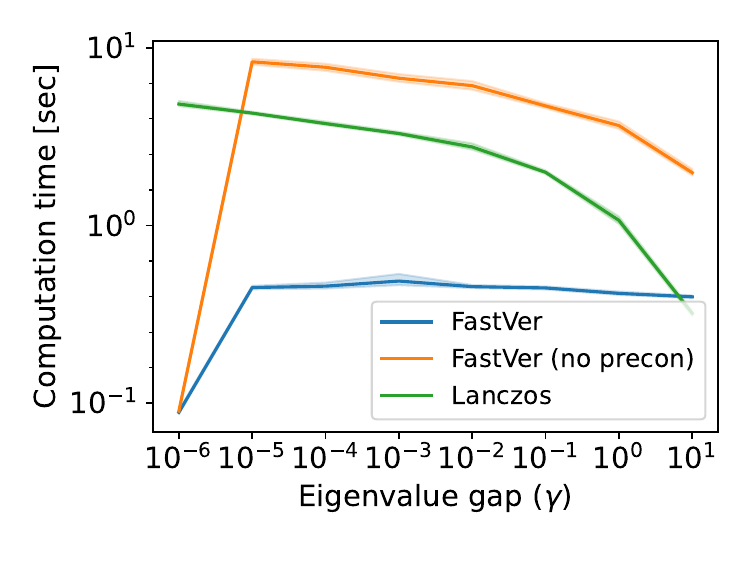}\label{verification_time_vs_gap_subfig}}
\subfigure[LOBPCG iterations vs.\ eigengap $\gamma$]{\includegraphics[width=.495\columnwidth]{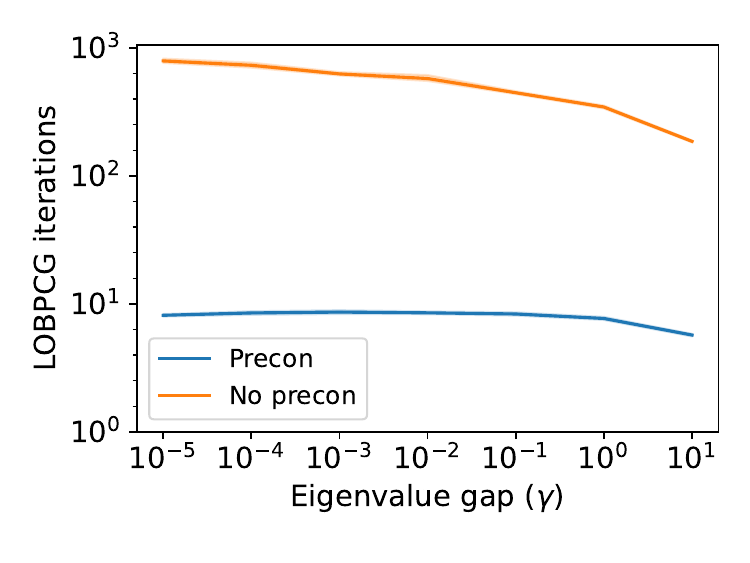}\label{num_iters_vs_gap_subfig}} \\
\subfigure[Computation time vs. size $N$]{\includegraphics[width=.495\columnwidth]{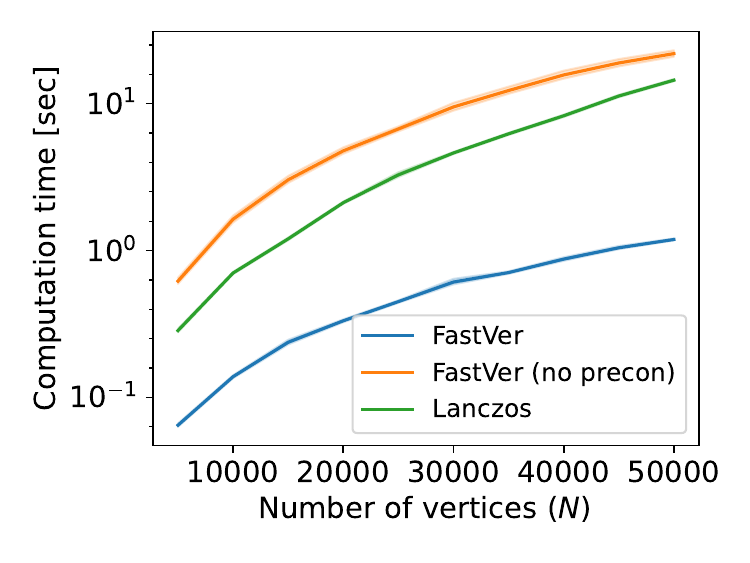}\label{verification_time_vs_size_subfig}}
\subfigure[LOBPCG iterations vs. size $N$]{\includegraphics[width=.495\columnwidth]{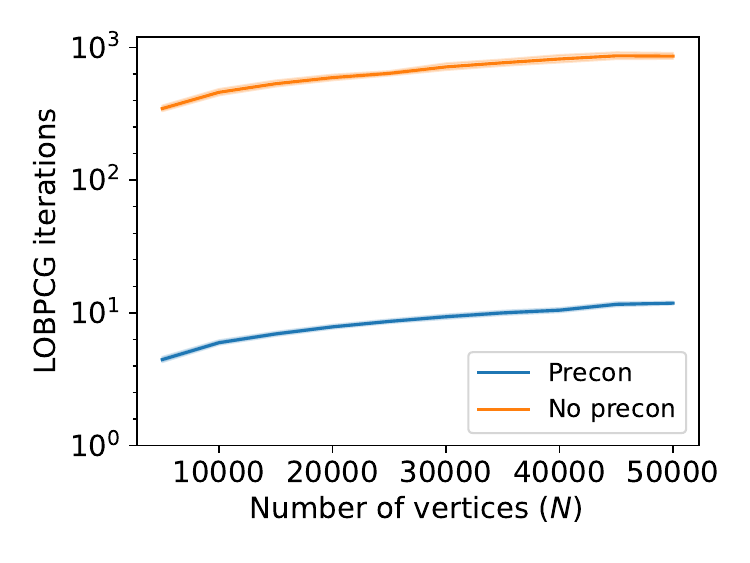}\label{num_iters_vs_size_subfig}}

\caption{Results for the simulation experiments.  Figs.\ \subref{verification_time_vs_gap_subfig} and \subref{verification_time_vs_size_subfig} plot the means and 95\% confidence interval shadings for the elapsed computation times of our proposed fast verification method (both with and without preconditioning) and the Lanczos method applied to 50 random realizations of the verification problem sampled from Algorithm \ref{test_matrix_generation_algorithm} while varying the eigenvalue gap $\gamma$ and problem size $N$ (respectively).  Figs. \subref{num_iters_vs_gap_subfig} and \subref{num_iters_vs_size_subfig} plot the means and 95\% confidence interval shadings for the number of LOBPCG iterations required in Algorithm \ref{FastVerification_algorithm} with and without preconditioning. \label{Synthetic_experiments_fig} }  
\end{figure}

In our first set of experiments, we study the effect of varying the eigenvalue gap $\gamma$.  We sample 50 matrices $\cert$ from Algorithm \ref{test_matrix_generation_algorithm} for each value of $\gamma = 10^k$, where $k \in \lbrace -6, \dotsc,  1 \rbrace$ (holding $N = 25000$), and then solve the verification problem (Problem \ref{verification_problem}) using  Algorithm \ref{FastVerification_algorithm} (both with and without the preconditioner $T$ described in Sec.\ \ref{modified_incomplete_symmetric_LDL_preconditioning}) and the Lanczos method.  We record the elapsed computation time for each method, as well as the number of iterations required by LOBPCG in each version of Algorithm \ref{FastVerification_algorithm}.  Results from this experiment are shown in Figs.\ \ref{verification_time_vs_gap_subfig} and \ref{num_iters_vs_gap_subfig}.

These results demonstrate that using LOBPCG together with the preconditioner proposed in Sec.\ \ref{modified_incomplete_symmetric_LDL_preconditioning} renders Algorithm \ref{FastVerification_algorithm} essentially insensitive to the eigenvalue gap $\gamma$ (note that the blue curves in Figs.~\ref{verification_time_vs_gap_subfig} and \ref{num_iters_vs_gap_subfig} are effectively flat).  In fact, we observe that Alg.~\ref{FastVerification_algorithm} actually achieves a significant \emph{gain} in performance for $\gamma \le \eta$, since in that case we can immediately detect $\cert$'s (numerical) positive-semidefiniteness using the test in lines \ref{M_Cholesky_factorization}--\ref{end_of_direct_test_for_positive_semidefiniteness}.  This small-$\gamma$ regime is especially important  because (as discussed in Sec.~\ref{related_work_section}) it is the most challenging for standard iterative eigenvalue methods, and in the context of the Burer-Monteiro approach is precisely the case that corresponds to certifying an \emph{optimal solution}.

In our second set of experiments, we study the effect of varying the problem size $N$.  As before, we sample 50 realizations of $\cert$ from Algorithm \ref{test_matrix_generation_algorithm} for each value of $N = 5000k$, where $k \in \lbrace 1, \dotsc, 10 \rbrace$, and then solve the verification problem (Problem \ref{verification_problem}) using  Algorithm \ref{FastVerification_algorithm} (again both with and without preconditioning) and the Lanczos method.  Results from this experiment are shown in Figs.\ \ref{verification_time_vs_size_subfig} and \ref{num_iters_vs_size_subfig}.  

These results show that the computational cost of Algorithm \ref{FastVerification_algorithm} (as measured in both elapsed time and LOBPCG iterations) scales gracefully with the size of the verification problem.  In particular, Fig.~\ref{verification_time_vs_size_subfig} shows that Alg.\ \ref{FastVerification_algorithm}'s running time increases approximately \emph{linearly} with  $N$ (note the logarithmic axis), which is consistent with the (approximately linearly) increasing cost of performing sparse matrix-vector multiplications and triangular solves with the operators $A$ and $T$.

\subsection{SLAM benchmarks}

\begin{figure}
 \subfigure[Computational cost of the Burer-Monteiro method]{\includegraphics[width=.49\columnwidth]{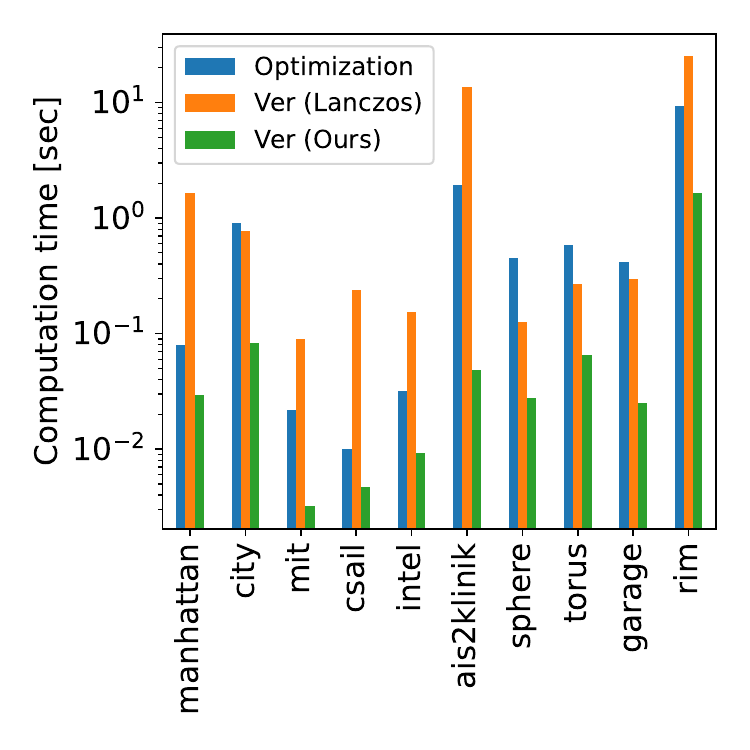}\label{SLAM_benchmarks_timing_subfig}}
\subfigure[Relative speed using Algorithm~\ref{FastVerification_algorithm} versus spectrally-shifted Lanczos]{\includegraphics[width=.49\columnwidth]{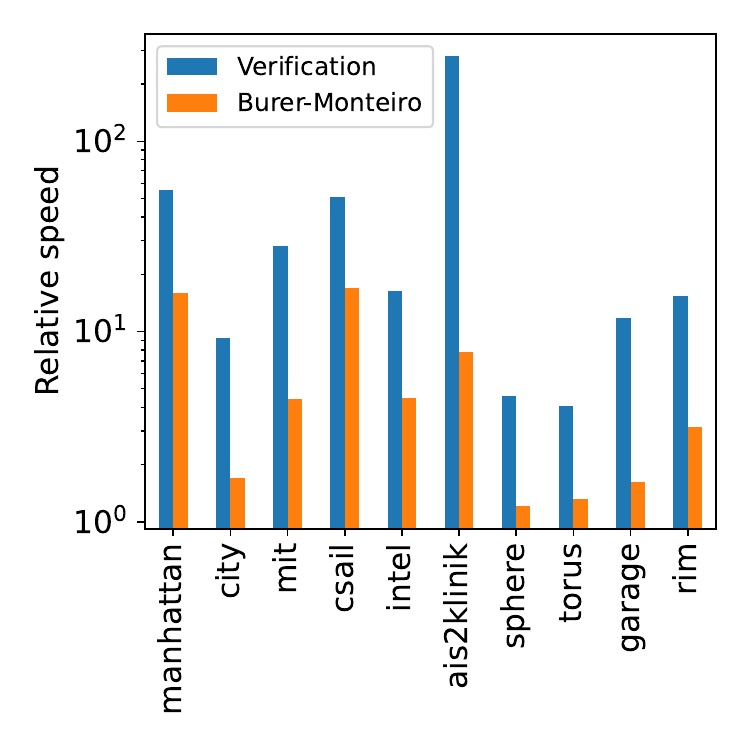}\label{SLAM_benchmarks_relative_speed_subfig}} \\
\caption{Results for the large-scale SLAM benchmarks.  \subref{SLAM_benchmarks_timing_subfig}:  The total computation time required to perform local optimization \eqref{low_rank_program} in the Burer-Monteiro method and solve the verification problem (Problem \ref{verification_problem}) using (i) the spectrally-shifted Lanczos method proposed in  \cite{Rosen2017Computational} and (ii) our proposed fast verification procedure (Algorithm \ref{FastVerification_algorithm}).  \subref{SLAM_benchmarks_relative_speed_subfig}:  The relative gain in computational speed achieved for both verification and the overall Burer-Monteiro method using our proposed fast verification scheme (Algorithm \ref{FastVerification_algorithm}) versus the spectrally-shifted Lanczos approach of \cite{Rosen2017Computational}.  }  
\label{SLAM_benchmarks_figure}
\end{figure}

In this experiment we evaluate the impact of our fast verification approach (Algorithm \ref{FastVerification_algorithm}) on the total computational cost of the Burer-Monteiro method.  To do so, we solve a sequence of large-scale semidefinite relaxations obtained from a standard suite of SLAM benchmarks (see \cite{Rosen2019SESync} for details).  For each problem instance, we record the total time required to perform the local nonlinear optimizations \eqref{low_rank_program}, as well as the total time required to solve the subsequent verification problems (Problem \ref{verification_problem}) using (i) our proposed fast verification method (Alg.~\ref{FastVerification_algorithm}) and (ii) the spectrally-shifted Lanczos approach proposed in \cite[Sec.\ III-C]{Rosen2017Computational}.  Results of this experiment are shown in Fig.\ \ref{SLAM_benchmarks_figure}.

These results clearly show that the verification step is the dominant cost of the Burer-Monteiro approach when using an unpreconditioned Krylov-subspace method, often requiring an order of magnitude more computational effort than nonlinear optimization.  Our fast verification procedure significantly accelerates this rate-limiting step in all tested cases, achieving a relative gain in speed of between \edit{4.08} (on \edit{\texttt{torus}}) and \edit{280} (on \edit{\texttt{ais2klinik}}) for the verification problems, and a corresponding gain of between \edit{1.21} (on \edit{\texttt{sphere}}) and \edit{16.9} (on \edit{\texttt{csail}}) for the end-to-end Burer-Monteiro method.  Crucially,  solution verification is no longer the dominant cost when using our fast verification scheme [note that the blue bars are uniformly taller than the green in Fig.\ \ref{SLAM_benchmarks_timing_subfig}].

\section{Conclusion}

In this letter we showed how to significantly accelerate solution verification in the Burer-Monteiro method, the rate-limiting step in many state-of-the-art certifiable perception algorithms.  We first established that the certificate matrices arising in the verification problem generically possess spectra that make it expensive to solve using standard iterative eigenvalue methods.  Next, we showed how to overcome this challenge using \emph{preconditioned} eigensolvers; specifically, we proposed a specialized solution verification algorithm based upon applying LOBPCG together with a simple yet highly effective incomplete factorization-based preconditioner.  Experimental evaluation confirms that our proposed verification scheme is very effective in practice, accelerating solution verification by  up to \edit{280x}, and the overall Burer-Monteiro method by up to \edit{16x}, versus the Lanczos method when applied to relaxations derived from large-scale SLAM benchmarks.

\bibliographystyle{IEEEtran}
\bibliography{references}

\end{document}